\documentclass[preprint,12pt, letterpaper]{elsarticle}

\usepackage{amssymb}
\usepackage{rotate}
\usepackage{setspace}
\usepackage{graphicx}
\usepackage[T1]{fontenc}
\usepackage{fancyhdr}
\usepackage{afterpage}
\usepackage{mathptmx}      
\usepackage{mathrsfs}
\usepackage{amsmath}
\usepackage{latexsym}

\newtheorem{proposition}{\textbf{Proposition}}[section]
\newtheorem{definition}{\textbf{Definition}}[section]
\newdefinition{remark}{Remark}[section]
\newtheorem{example}{\textbf{Example}}[section]

\newproof{proof}{\textbf{Proof}}
\newproof{pot}{Proof of Theorem \ref{thm2}}

\journal{ArXiv}

\begin{document}
\begin{spacing}{1.0}
\begin{frontmatter}

\title{Redefinition of the concept of fuzzy set based on vague partition from the perspective of axiomatization}

\renewcommand{\thefootnote}{\fnsymbol{footnote}}
\author{Xiaodong Pan\footnote{Corresponding author.\\
E-mail address: xdpan1@163.com(Xiaodong Pan)}, Yang Xu}

\address{School of Mathematics, Southwest Jiaotong University, West Section, High-tech Zone, Chengdu, Sichuan, 611756, P.R. China}

\begin{abstract}
Based on the in-depth analysis of the essence and features of vague phenomena, this paper focuses on establishing the axiomatical foundation of membership degree theory for vague phenomena, presents an axiomatic system to govern membership degrees and their interconnections. On this basis, the concept of vague partition is introduced, further, the concept of fuzzy set introduced by Zadeh in 1965 is redefined based on vague partition from the perspective of axiomatization. The thesis defended in this paper is that the relationship among vague attribute values should be the starting point to recognize and model vague phenomena from a quantitative view.
\end{abstract}


\end{frontmatter}


\section{The mathematical analysis of vagueness}
Fuzzy sets introduced by Zadeh \cite{Zadeh1965} in 1965 are designed to model or describe the extensions of vague concepts, such as \textit{Young, Warm, Tall, etc.}. To be more precise, the extension of a vague concept $\varphi$ is taken to be a fuzzy set $A$, which is defined by a membership function $\mu_{\varphi}: U \rightarrow [0,1]$, where $U$ is the domain of discourse. For every element $x \in U$, the value $\mu_{\varphi}(x) \in [0,1]$ is the membership degree of $x$ in $A$, or the degree to which one believes that the element $x$ have the attribute $\varphi$. And after that, some important contributions are made to fuzzy sets and fuzzy logic  by Zadeh and others \cite{Bed13, Cin11, DDH2015, Dubois99, Haj98, Nov99, Nov06, Nov12, Pav79, Zadeh2015}.
\par However, is fuzzy set or fuzzy logic the suitable and effective tool for dealing with vagueness? Seemingly the answer is not so obviously. In \cite{Beh09}, B\v{e}hounek said that "\textit{Fuzzy logic cannot claim to be \emph{the} logic of vagueness $\cdots$ most of which are not captured by deductive fuzzy logic}". Nov\'{a}k said in \cite{Nov12} that "\textit{fuzzy logic is not the logic of vagueness but the logic of ordered structure $\cdots$ there are not yet many results on the applied side $\cdots$ It is still not fully clear which logic is the most convenient to solve problems related to models of vagueness and their applications}". From these comments, you can see that fuzzy set theory is still at its first stage, its theoretical foundation is far from complete.
\par In the following, we list three typical questions in Zadeh's fuzzy set theory:
\par First, for a given vague concept, how to define the membership function of its extension? And how to justify its rationality? For example, for vague concept "tall", if we define its membership functions on the domain of discourse "Cambodian men" and "Dutch men" respectively, are these two functions same? Another example, if we define the membership functions for "tall" on the domain of discourse "Japanese men" in 1900 and in 2016 respectively, are these two functions same? Common sense tells us that these functions should be very different. On the other hand, whether any mapping from any nonempty set $U$ to $[0, 1]$ (e.g. $y = \sin(1/x)$ or any probability distribution function) will determine a fuzzy set? If the answer is yes, then which vague concept will be modelled by the mapping? In addition, given a membership degree $a \in [0,1]$ with respect to a fuzzy set $A$, how to find the objects whose membership degree is exactly equal to $a$? or more general, is approximately equal to $a$? These questions are still hard  to answer in Zadeh's fuzzy set theory. Intuitively, the time and space factors should be considered when we deal with vagueness.
\par Second, suppose the time and space have been fixed when we discuss a vague concept (e.g. "young", denoted by $\varphi$), whether we should consider the relationship between this vague concept "young" and its related vague concepts (e.g. "old", denoted by $\psi$) in defining the membership function $\mu_{\varphi}$ of its extension? To be more precise, the membership functions $\mu_{\varphi}$ and  $\mu_{\psi}$ are independent of each other, or are associated with each other? If they are independent of each other, then is it reasonable to define $\mu_{\varphi}(x) = \mu_{\psi}(x) = 0.7$  for an object $x$? If they are correlative of each other, then how to reflect their relationship into the definitions of their membership functions?
\par Third, the membership function $\mu_{A}$ of a fuzzy set $A$ is a mapping from the domain of discourse $U$ to $[0,1]$, where $U$ is the collection of objects. Under this definition, whether the relations between fuzzy sets in the same domain of discourse $U$ should always be truth-functional? Its answer is still not clear. In \cite{Smi97}, D. Edgington raised doubts about this. She presented the following example: suppose we have a collection $U$ of balls of various sizes and colors. These are independent variables: how close a ball is to a clear case of "\textit{small}" (in the context) is unaffected by its color; and how close it is to a clear case of "\textit{red}" is unaffected by its size. Let $R$ and $S$ be the fuzzy sets in $U$, denotes vague concepts "\textit{red}" and "\textit{small}", respectively.  For balls $a, b$ and $c$, suppose: $$R(a) = 1, R(b) = R(c) = 0.5;\ S(a) = S(b) = 0.5, S(c) = 0.$$ According to the operations "$\cap$(intersection)" and "$\cup$(union)" for fuzzy sets, $(R \cap S)(a) = \min\{R(a), S(a)\} = 0.5 =  \min\{R(b), S(b)\} = (R \cap S)(b)$,  $(R \cup S)(b) = \max\{R(b), S(b)\} = 0.5 =  \max\{R(c), S(c)\} = (R \cup S)(c)$. But it is plausible that $a$ is a better case for vague concept "\textit{red and small}" than $b$, and $b$ is a better case for "\textit{red or small}" than $c$. What's the reasons? Maybe someone will say that we can use \textit{t}-norms and \textit{t}-conorms (or other operators) to model the intersection and union of fuzzy sets, but which one will work? In our view, the reasons lie not in the interpretations of set-theoretic operations.
\par This paper aims at doing some attempts  to answer the above three questions. For simplicity, in this paper, we will focus on our discussion of vague concepts in a fixed time and space, and believe that all things are interrelated and interact on each other, we have to learn to contradistinguish one thing from another. For example, the existence of other colors except for "\textit{red}" makes it possible for us to distinguish "\textit{red}"  from other colors, thus if you want to understand what "\textit{red}" is, then you need also to know what "\textit{yellow}", "\textit{orange}" and other colors are. For the same reason, if you want to understand what a tall man is like, you also need to know what a short man is like. Inspired by this thought, the relationship between a vague predicate and its other closely related vague predicates will be the starting point of our discussion.  From a global and overall of view, we will try to establish a mathematical model of treating phenomena of vagueness from an axiomatic point of view, it is our hope that the model could serve as the theoretical foundation of fuzzy set theory and its application, and further serve as the starting point of formalized theory of dealing with vague phenomena.
\par As a preliminary, firstly we must know what is vagueness, and what is its main characteristics? Otherwise, our study will be like that water without a source and a tree without roots.
\par In \cite{Zadeh1978}, Zadeh wrote "\textit{Although
the terms fuzzy and vague are frequently used interchangeably in the literature, there is, in fact, a significant difference between them. Specifically, a proposition, $p$, is fuzzy if it contains words which are labels of fuzzy sets; and $p$ is vague if it is both fuzzy and insufficiently specific for a particular purpose.}" However, what words should be the labels of fuzzy sets, Zadeh didn't  given further explanation.
\par In this paper, the terms "fuzzy" and "vague" will be used interchangeably unless otherwise stated, its exact meaning will be explained in detail in next section.

\section{What is vagueness?}

In the 1902 Dictionary of Philosophy and Psychology \cite{Peirce02}, Charles Sander Peirce explained the entry "vague" as follows:
\par \textit{A proposition is vague when there are possible states of things concerning which it is intrinsically uncertain whether, had they been contemplated by the speaker, he would have regarded them as excluded or allowed by the proposition. By intrinsically uncertain we mean not uncertain in consequence of any ignorance of the interpreter, but because the speaker's habits of language were indeterminate. (Peirce 1902, 748)}
\par  From Peirce's description, vagueness is intrinsically indeterminacy, which means a certain state of object that is not absolutely affirmative or negative to say which attribute should be assigned to the object, and vagueness is related closely to the use of natural language, and will result in borderline cases. But, how exactly does vagueness arise? what exactly vague phenomena are?  And what is its essence?
\subsection{Sorites Paradoxes}
The word "sorites" derives from the Greek word for heap, the paradox is so named because of its original characterization, attributed to Eubulides of Miletus. Vague predicates are susceptible to Sorites Paradoxes. As we mentioned above, the extension of a vague concept (or predicate) has borderline cases, can not be clear and well defined. In other words, there are objects which one cannot say with certainty whether belong to a group of objects which are identified with this concept or which exhibit characteristics that have this predicate. Hence, the "class" $X$ of all objects which have the property $\varphi$ (vague predicate) cannot be taken as a set (crisp set) since the vagueness of property $\varphi$ makes it impossible for us to characterize the "class" precisely and unambiguously.  For example, the predicate "\textit{is tall}" is vague because a man who is 1.75 meters in height is neither clearly tall nor clearly non-tall. No amount of conceptual analysis or empirical investigation can determine whether a 1.75-meter man is tall, as we mentioned above, it is a kind of intrinsically indeterminacy.
\par Many people will agree that almost all vague phenomena can be attributed to the vagueness of predicates. Consider the predicate "\textit{is a heap}", the paradox goes as follows: consider a heap of wheat from which grains are individually removed. One might construct the argument, using premises, as follows:
\begin{itemize}
\item Premise 1: $10^{10}$ grains of wheat is a heap of wheat;
\item Premise 2: when we remove one grain of wheat from a heap of wheat, the remain is still a heap.
\end{itemize}
Repeated applications of Premise 2 (each time starting with one fewer grain) eventually forces one to accept the conclusion that a heap may be composed of just one grain of wheat. There are some other variations of the paradox, in fact, this paradox can be reconstructed for a variety of predicates, for example, with "\textit{is tall}", "\textit{is rich}", "\textit{is old}", "\textit{is blue}", "\textit{is bald}", and so on.
\par The root cause of the paradox lies in the existence of borderline cases brought about by the use of vague predicates which are usually expressed by natural language. Even if you may know all relevant information about vague predicates, such as, the exact number of these wheats, the precise value of the height of a man, the precise value of the wavelength of a kind of color, etc., you are still unable to determine whether these wheats form a heap, whether this man is a tall man, whether this kind of color is red, all these are not because we can't understand these predicates, but because of the vagueness of these predicates. That is to say, the vagueness of a vague predicate doesn't come from the shortcomings of the cognitive abilities of the human itself, but is the result of combined effect by a kind of objective attribute existing in the objects itself and human subjective cognitive style.

\subsection{Multidimensional vagueness}
So far we have considered only these vague predicates which are determined by an one-dimensional variation (or only one attribute), such as height for "\textit{tall}", age for "\textit{young}" and temperature for "\textit{hot}". But in many real problems, the vague predicates are multidimensional: several different dimensions of variation (or several attributes) are involved in determining their applicability. The applicability of "\textit{big}", when it is used to describe a man, depends on both height and volume. Whether a ball is counted as a "\textit{small and red ball}" depends not only on the volume of ball but also on its color. Moreover, there are still some vague predicates in which it is not even a clear-cut set of dimensions determining the applicability of these predicates: it is not clear which factors are related to each other and blend into one another. For example, the applicability of "\textit{good}", which is used to describe a student, whether a student should be counted as "\textit{good}", maybe different people have different views on it, it is very difficult to present a universally accepted criterion to judge whether a student is good. Of course, multidimensional vague predicates also share all these features mentioned above of vague predicates.
\par It is important to note, however, that when one applies fuzzy sets to model multidimensional vagueness, then the range of membership functions should be a subset in $[0,1]^{n}$ instead of $[0,1]$, where $n$ denotes the number of dimensions of a multidimensional vague predicate. In this way, it is easy to remove D. Edgington's doubts mentioned in Section 1. More generally, one can take a complete distributive residuated lattice \cite{War38} as the range of membership function. More details please refer to \cite{Gog67, Gog69}.
\subsection{Higher-order vagueness}
Let $F$ be a vague predicate, then there is no (sharp) boundary between the objects that determinately satisfy $F$ and those that do not determinately satisfy $F$, that is, it admits borderline case where it is unclear whether $F$ applies. This is the so-called first-order vagueness. Intuitively, it seems that there is also no (sharp) boundary between the objects that determinately satisfy $F$ and the borderline objects in the borderline case, nor is there a sharp boundary between the borderline objects and those that do not determinately satisfy $F$. This is called as second-order vagueness. Proceeding in this way, one can define the concepts of "third-order vagueness", "fourth-order vagueness" etc., which are uniformly referred to as "higher-order vagueness".
\par When one applies fuzzy sets to model higher-order vagueness, the problem is how to define the membership functions of fuzzy sets. For example, let $\varphi$ be the statement "\textit{Tom is tall, if Tom is 1.8 metres in height}", how much is the truth value of $\varphi$? If one set its truth value equal to 0.6 (or other any definite value), then you might ask: "why not 0.61 or 0.59?" Meanwhile, another question also arises: if we consider that the truth value of "\textit{John is tall}" is bigger than 0.6, then how much John's height should be? All these are the reflection of higher-order vagueness in fuzzy set theory.
\par In fuzzy set theory, several solutions have been proposed by Zadeh \cite{Zad75-1}, Grattan-Guiness \cite{Gra75} and others \cite{Gor87, Hu14, HuW14, Kar01, Men07, Sad13, Zadeh1975} in response to the above problems. Zadeh took linguistic terms as the grades of membership of fuzzy sets, and then these linguistic terms were modeled by fuzzy sets of type-$n$ whose membership function ranges over fuzzy sets of type $n - 1$. The membership function of a fuzzy set of type-1 ranges over the interval $[0, 1]$. Another way to deal with higher vagueness in fuzzy set theory is to replaced the interval $[0, 1]$ by the set of subintervals of $[0, 1]$, which has been proposed by Grattan-Guiness. I. \cite{Gra75} in 1975. However, the above two strategies can only alleviated the problem of higher vagueness to some extent, but it can not solve the problem authentically. In fact, for fuzzy sets of type-$n$, we still need to define the membership functions of fuzzy sets of type $n - 1$. For interval valued fuzzy sets, we still need to define the subintervals as the grades of membership of fuzzy sets. Hence, the difficulties here is similar to that in Zadeh's fuzzy sets.

\subsection{Vagueness vs. Uncertainty, Randomness}
Vagueness is also often confused with uncertainty and randomness. In general, uncertainty is applied to predictions of future events, to physical measurements that are already made, or to the unknown. It arises in partially observable and/or stochastic environments, as well as due to ignorance and/or indolence. Douglas W. Hubbard \cite{Hubbard2010} defined uncertainty as: \textit{"The lack of complete certainty, that is, the existence of more than one possibility. The "true" outcome/state/result/value is not known"}. Uncertainty is encountered when an experiment (process, test, etc.) is to proceed, the result of which is not known to us. Hence, uncertainty is always connected with the question whether the given event may be regarded within some time period, or not; there is no uncertainty after the experiment was realized and the result is known to us.
\par Uncertainty emerges probably due to the lack of enough knowledge, probably due to the shortcomings of our cognitive abilities, and also probably due to the relatively poor technical conditions etc. Frequently, uncertainty will disappear as long as these situations have been improved. Here you see that uncertainty differs from vagueness, the latter has nothing to do with these outside conditions. Vagueness only concerns the object itself under consideration and the way how it is delineated according to its certain attribute, and won't disappear as time passed. You can say that the difference between uncertainty and vagueness corresponds to the difference between potentiality and factuality.
\par In addition, Randomness is a specific form of uncertainty, and thus differs from vagueness. Randomness suggests a non-order or non-coherence in a sequence of symbols or steps, such that there is no intelligible pattern or combination. Probability theory is the mathematical analysis of random phenomena, probability can be thought of as a numerical measure of the likelihood that an event will occur, its value is a number between 0 (0 percent chance or will not happen) and 1 (100 percent chance or will happen). In this sense, probability is similar to membership degrees in fuzzy set theory.

\subsection{What is vagueness?}
Now, let's go back to the previous questions: how exactly does vagueness arise? what exactly fuzzy phenomena are?  And what is its essence?
\par Based on the previous discussion, the thesis about vagueness defended in this paper is that vagueness arises in the process of classifying objects, it is a kind of manifestation of the continuity and gradualness existing in the process of development and evolution of objects, and is the result of combined effect by the continuity and gradualnes existing in the objects itself and human subjective cognitive style. Vague phenomena are the external manifestations of vagueness.
\par For example, a person is impossible to become an adult from a baby in an instant, the process is continual and evolutionary. When one regards this process in a discrete point of view based on certain specific purpose (classification), or divides this process into several discrete stages in a special way (describe each stage in a vague predicate), consequently the continuity and gradualness existing in objects themselves can not be revealed completely by such partition way, and then the vagueness arises.
\par An usual way to discretize a continual and evolutionary process is to delineate the process by using some natural linguistic terms (words or phrases, i.e., vague predicates), a natural linguistic term labels a stage in the process, these natural linguistic terms denote the main characteristic of each corresponding stage respectively. In other words, natural language provides us with a very useful tool to describe a continuous process in a discrete way. Hence, vagueness usually arises together with the use of natural language.
\par Based on the above analysis, the occurrence of vagueness is not due to the lack of human cognitive ability and technological means, and is also not due to the complexity of objective things themselves, but is due to a specific cognitive style (discretizing the continual and evolutionary process of objects) for objective things based on a certain cognitive aim (classifying objects). In this sense, vagueness is a bridge to link discretization and continuity.
\par We make further explanation of vagueness by the following example. Let the set of all Chinese be the domain of discourse, and consider the attribute "\textit{Age}" of Chinese. According to our common sense, a person's age should be in the scope of 0 to 200 (an optimistic estimate). It is a continuous process for one's age to change from 0 to 200, if we partition the process into several stages by labelling each stage with one of the following natural linguistic terms \textit{Infant, Childish, Juvenile, Young, Adult, Middle age, Elderly, Old}, here the set $\Omega = \{$\textit{Infant, Childish, Juvenile, Young, Adult, Middle age, Elderly, Old}$\}$ will be called as a vague partition of Chinese with respect to the attribute "\textit{Age}", then this discretized cognitive style (the set $\Omega$) hides the inherent continuity and gradualness existing in the attribute "\textit{Age}". Consequently, these predicates such as "\textit{is young}", "\textit{is adult}", "\textit{is old}" etc., show vagueness. 
\par According to the origins, fuzzy phenomena can be divided into different types: one-dimensional fuzzy phenomena, two-dimensional fuzzy phenomena, etc.

\section{Some basic terminologies}
In this section, we define several basic notions that will be needed in later sections: elementary vague attribute, vague attribute, vague space and vague judgement.
\par In general, a concept (its meaning is clear for us) is determined by or involved with one or a group of attributes, these attributes can be one-dimensional, and can also be multidimensional. For example, \textit{Weight, Height, Age} are one-dimensional attributes with respect to the concept "Man", \textit{Length, Width, Height, Color} are also one-dimensional attributes with respect to the concept "Box", but \textit{Area, Volume} are multidimensional attributes with respect to the concept "Room". It is worth noting that a multidimensional attribute can be determined usually by several one-dimensional attributes. For example, $Area$ is determined by one-dimensional attributes \textit{Length} and \textit{Width}, and can be regarded as the Cartesian product $Length \times Width$.
\par Let $C$ be a concept, we need to make a finite (or countable) partition of its extension (which is usual a crisp (or classical) set) by linguistic terms according to one or a group of attributes of $C$. From the previous exposition, we will find that the partition is not sharply if the attributes acting as the standard of the partition is characterized by continuity, in this case, we call such partition as vague partition \cite{Pan15}. In this paper, we only consider the vague partition according to certain one-dimensional attribute.
\par As an example, when we speak of someone's height, we might prefer to say that he (or she) is \textit{tall}, \textit{short}, or \textit{medium} etc., rather than to say that he (or she) is 1.753 metres in height. Let $C$ be the concept "Man" and $\varphi_{C}$ the one-dimensional attribute "\textit{Height}". If we need to divide all men which are in certain fixed time and space (e.g. Dutch men) into seven (of course, it can also be other numbers) classes, then the set whose elements consist of names of all these classes, like the following $$\{\text{Very Very Short, Very Short, Short, Medium, Tall, Very Tall, Very Very Tall}\}$$ is called the elementary set of vague attribute values of $\varphi_{C}$, denoted by $\Omega_{\varphi_{C}}$, then the set $\Omega_{\varphi_{C}}$ can be viewed as a vague or fuzzy partition of the extension of $C$ according to the attribute $\varphi_{C}$.  The mathematical definition of vague partition will be presented in the latter section. In what follows, we always denote the elementary set of vague attribute values of $\varphi_{C}$ with respect to the concept $C$ by $\Omega_{\varphi_{C}}$ (which can be effectively equivalent to the sample space in probability theory, the only difference is that in vague membership degree theory, the elementary set of vague attribute values is alterable according to different requirements of applications). The elements in $\Omega_{\varphi_{C}}$ are called elementary vague attribute values with respect to $\varphi_{C}$. In addition, the domain of discourse of the attribute $\varphi_{C}$, or the range of values of measurement of $\varphi_{C}$, is denoted by $U_{\varphi_{C}}$. Based on the analysis in Section 1, the set $U_{\varphi_{C}}$ should be an interval. For example, let $C$ be the concept "Man" and $\varphi_{C}$ the attribute "\textit{Age}", then you can take the universe $U_{\varphi_{C}}$ to be the interval $[0, 200]$ according to the common sense.
\par Let $\mathcal{F} = \{\bot, \top, \neg, \barwedge, \veebar\}$, where $\bot, \top$ represent  "nonexistent vague attribute value" (e.g.  "$x$ is $\bot$" means that $x$ has no a certain attribute.) and  "intrinsic vague attribute value" (e.g.  "$x$ is $\top$" means that $x$ must have a certain attribute.) respectively, and $\neg, \barwedge$ and $\veebar$ represent the connectives "negation", "and" and "or" respectively, which are used as the operations among various vague attribute values. The set of vague attribute values of the concept $C$ with respect to the attribute $\varphi_{C}$, denoted by $\Sigma_{\varphi_{C}}$, is the smallest set such that
\begin{itemize}
\item $\bot, \top \in \Sigma_{\varphi_{C}}$.
\item $\Omega_{\psi_{C}} \subset \Sigma_{\varphi_{C}}$.
\item If $A \in \Sigma_{\varphi_{C}}$, then $\neg A \in \Sigma_{\varphi_{C}}$.
\item If $A, B \in \Sigma_{\varphi_{C}}$, then $A \barwedge B, A \veebar B \in \Sigma_{\varphi_{C}}$.
\item If $A_{1}, A_{2}, \cdots, A_{n}, \cdots \in \Sigma_{\varphi_{C}}$, then $\veebar_{i = 1}^{\infty}A_{i}, \barwedge_{i = 1}^{\infty}A_{i} \in \Sigma_{\varphi_{C}}$.
\end{itemize}
In what follows, we call the triple $(\Omega_{\varphi_{C}}, \Sigma_{\varphi_{C}}, \mathcal{F})$ vague space with respect to $\varphi_{C}$ (which can be effectively equivalent to the event field in probability theory), and call these elements in $\Sigma_{\varphi_{C}}$ vague attribute values with respect to $\varphi_{C}$. You may have noticed that a vague attribute value can be in some sense regarded as an evaluating linguistic expression which has been introduced by Dvo\v{r}\'{a}k and Nov\'{a}k in \cite{Dvorak04}. As mentioned above, $Short$ is an elementary vague attribute value with respect to "\textit{Height}", the element $Short \veebar Medium$ is also a vague attribute value with respect to "\textit{Height}", which means "\textit{Short or Medium}", and can be named as the attribute "\textit{Lower Medium}".
\par Let $x \in U_{\varphi_{C}}$, a vague judgement in vague space $(\Omega_{\varphi_{C}}, \Sigma_{\varphi_{C}}, \mathcal{F})$ with respect to $x$ is a procedure to determine these degrees to which these vague attribute values in $\Omega_{\varphi_{C}}$ are possessed by the element $x$ respectively. Put it in another way, a vague judgement in vague space $(\Omega_{\varphi_{C}}, \Sigma_{\varphi_{C}}, \mathcal{F})$ with respect to $x$ is a series of statements (vague propositions) "$x$ is $p$", where $p \in \Omega_{\varphi_{C}}$. For instance, let $C$ be the concept "Man" and $\varphi_{C}$ the attribute "\textit{Height}". Define $\Omega_{\varphi_{C}}$ and $U_{\varphi_{C}}$ as $$\{\text{Very Very Short, Very Short, Short, Medium, Tall, Very Tall, Very Very Tall}\}$$ and $[0, 3]$, respectively, and $\mathcal{F} = \{\bot, \top, \neg, \barwedge, \veebar\}$ and $\Sigma_{\varphi_{C}}$ is given accordingly. If $x = 1.7 \ \text{metres} \in U_{\varphi_{C}}$, then a vague judgement in vague space $(\Omega_{\varphi_{C}}, \Sigma_{\varphi_{C}}, \mathcal{F})$ with respect to $x$ consists of the following statements: "$x$ \textit{is very very short}"; "$x$ \textit{is very short}"; "$x$ \textit{is short}"; "$x$\textit{ is medium}"; "$x$\textit{ is tall}"; "$x$ \textit{is very tall}"; "$x$\textit{ is very very tall}". Furthermore, we can assign a real number in the unit interval $[0, 1]$ to each of these statements, which are referred to as the truth values of these statements. The truth value of the statement "$x$ is $p$" ($p \in \Omega_{\varphi_{C}}$) is a measure or estimation of the extent that one trusts that $x$ is $p$. According to fuzzy set theory, this truth value is also called membership degree of $x$ in fuzzy set $A$, where $A$ is determined by the vague attribute value $p$, and is a mathematical model of the extension of $p$. In a vague judgement, the relationship among membership degrees of vague propositions is the key point this paper focuses on, we will discuss this topic in detail in the next section.
\par Now, let's consider the previous example. "\textit{tall}" (denoted by $A$) is a vague attribute value of attribute "\textit{Height}", $A \in \Omega_{\text{Height}}$. We need to define the membership function $\mu_{A}$ to model the extension of "\textit{tall man}". Suppose that Tom is a 1.75 metres tall man, then how much the number $\mu_{A}(\text{Tom})$ should be?
\par Let's assume that Tom is an American, the time is 2015. We will find that the value $\mu_{A}(\text{Tom})$ will depend on the set $\Omega_{\text{Height}}$. If $$\Omega_{\text{Height}} = \{\text{short, medium height, tall}\},$$ that is, people are divided into three classes, then you can take $\mu_{A}(\text{Tom}) = 0$. This is because in this case, you really believe that Tom is of medium build (because the average height of the American men is 1.75 metres), the degree of truth that Tom is of medium build will be 1, and so $\mu_{A}(\text{Tom})$ should be zero. But if $\Omega_{\text{Height}} = \{\text{short, tall}\}$, that is, people are divided into only two classes. In this case, Tom will be incorporated into either the class of "\textit{short men}" or the class of "\textit{tall men}", then maybe $\mu_{A}(\text{Tom}) = 0.6$ will be reasonable.
\par This example tells us the fact that when we model a vague predicate using a numerical method,  it involves vague partition, that is, the set $\Omega_{\varphi_{C}}$, must be considered.
\section{The axioms for membership degrees}
In this section, we present the system of axioms that will govern the relations among various degrees of membership. Before that, we first review several fundamental concepts, "Triangular norm", "Triangular conorm" and "Strong negation", which will be needed for the rest part of this paper. For more details, please refer to \cite{Klement}.
\begin{definition}
A triangular norm (t-norm for short) is a binary operation $T$ on the unit interval $[0,1]$, i.e., a function $T: [0, 1]^{2} \rightarrow [0, 1]$, such that for all $x, y, z \in [0, 1]$, the following four axioms are satisfied:\\
(T1) \ $T(x, y) = T(y, x)$,\hspace*{\fill} (commutativity)\\
(T2) \ $T(x, T(y, z)) = T(T(x, y), z)$,\hspace*{\fill} (associativity)\\
(T3) \ $T(x, y) \leqslant T(x, z)$ whenever $y \leqslant z$,\hspace*{\fill} (monotonicity)\\
(T4) \ $T(x, 1) = x$.\hspace*{\fill} (boundary condition)
\end{definition}
\begin{example}
The following are the four basic t-norms $T_{M}, T_{P}, T_{L},$ and $T_{D}$ given by, respectively:\\
\par $T_{M}(x, y) = \min\{x, y\}$,\hspace*{\fill} (minimum)
\par $T_{P}(x, y) = x \cdot y$,\hspace*{\fill} (product)
\par $T_{L}(x, y) = \max\{x + y - 1, 0\}$,\hspace*{\fill} ({\L}ukasiewicz t-norm)
\par $T_{D}(x, y) = \left\{ \begin{array}{ll}
0, &\mbox{if \ $(x, y) \in [0, 1)^{2}$,} \\
\min\{x, y\}, &\mbox{otherwise.}
\end{array} \right. $  \hspace*{\fill} (drastic product)
\end{example}
\begin{definition}
A triangular conorm (t-conorm for short) is a binary operation $S$ on the unit interval $[0,1]$, i.e., a function $S: [0, 1]^{2} \rightarrow [0, 1]$, which, for all $x, y, z \in [0, 1]$, satisfies ($T1$)-($T3$) and \\
(S4) \ $S(x, 0) = x$. \hspace*{\fill} (boundary condition)
\end{definition}
\begin{example}
The following are the four basic t-conorms $S_{M}, S_{P}, S_{L},$ and $S_{D}$ given by, respectively:\\
\par $S_{M}(x, y) = \max\{x, y\}$,\hspace*{\fill} (maximum)
\par $S_{P}(x, y) = x + y - x \cdot y$,\hspace*{\fill} (probabilistic sum)
\par $S_{L}(x, y) = \min\{x + y, 1\}$,\hspace*{\fill} ({\L}ukasiewicz t-conorm, bounded sum)
\par $S_{D}(x, y) = \left\{ \begin{array}{ll}
1, &\mbox{if \ $(x, y) \in (0, 1]^{2}$,} \\
\max\{x, y\}, &\mbox{otherwise.}
\end{array} \right. $  \hspace*{\fill} (drastic sum)
\end{example}
\par Since both t-norm and t-conorm are algebraic operations on the unit interval $[0, 1]$, it is of course also acceptable to use infix notations like $x \otimes y$ and $x \oplus y$ instead of the prefix notations $T(x, y)$ and $S(x, y)$ respectively. In what follows, we will use these infix notations most of the time.
\begin{definition}
(i) A non-increasing function $N: [0, 1] \rightarrow [0, 1]$ is called a negation if
 \par (N1) $N(0) = 1$ and $N(1) = 0$.\\
(ii) A negation $N: [0, 1] \rightarrow [0, 1]$ is called a strict negation if, additionally,
 \par (N2) $N$ is continuous.
 \par (N3) $N$ is strictly decreasing.\\
(iii) A strict negation $N: [0, 1] \rightarrow [0, 1]$ is called a strong negation if it is an involution, i.e., if
 \par (N4) $N \circ N = id_{[0,1]}$ is continuous.
\end{definition}
\par It is obvious that $N: [0, 1] \rightarrow [0, 1]$ is a strict negation if and only if it is a strictly decreasing bijection.
\begin{example}
(i) The most important and most widely used strong negation is the standard negation $N_{S}: [0, 1] \rightarrow [0, 1]$ given by $N_{S} = 1 - x$. It can be proved that each strong negation can be seen as a transformation of the standard negation, see \cite{Klement}.\\
(ii) The negation $N: [0, 1] \rightarrow [0, 1]$ given by $N(x) = 1 - x^{2}$ is strict, but not strong.\\
(iii) An example of a negation which is not strict and, subsequently, not strong, is the G$\ddot{o}$del negation $N_{G}: [0, 1] \rightarrow [0, 1]$ given by
\[
N_{G}(x) = \left\{
\begin{array}{ll}
1, &\mbox{if \ $x = 0$,}\\
0, &\mbox{if \ $x \in (0, 1]$.}
\end{array}
\right.
\]
\end{example}
\par In order to deal with vagueness mathematically (numerically or formalized), and establish the rigorous foundation for the mathematical analysis of vagueness, we propose the following axioms by which the membership degrees and their interconnections are to be governed.
\par First of all, we only consider one-dimensional vague attribute values. Let $C$ be a concept, $x \in U_{\varphi_{C}}$ and $(\Omega_{\varphi_{C}}, \Sigma_{\varphi_{C}}, \mathcal{F})$ a vague space with respect to the attribute $\varphi_{C}$. The vague membership space with respect to $\varphi_{C}$ (which is one-dimensional) and associated with $x$ is a quintuple $(\Omega_{\varphi_{C}}, \Sigma_{\varphi_{C}}, \mathcal{M}_{x}, \mathcal{F}, \mathcal{T})$, where $\mathcal{T} = \{N, \oplus, \otimes\}$ consists of a t-norm $\otimes$, a t-conorm $\oplus$ and a strong negation $N$. $\mathcal{M}_{x}$ is a real value function (it is usually called the vague membership measure \cite{Pan16} on vague space $(\Omega_{\varphi_{C}}, \Sigma_{\varphi_{C}}, \mathcal{F})$, or membership measure for short) from $\Sigma_{\varphi_{C}}$ to $[0, 1]$ with respect to $x$, satisfies the following axioms:
\begin{itemize}
\item {\bf Axiom I.} For any $A \in \Sigma_{\varphi_{C}}$, $0 \leqslant \mathcal{M}_{x}(A) \leqslant 1$, and there is at least one element $p \in \Omega_{\varphi_{C}}$ such that $\mathcal{M}_{x}(p) > 0$.\\
 Axiom I says that for any $A \in \Sigma_{\varphi_{C}}$, the degree to which $x$ has the vague attribute value $A$ is between 0 and 1, and $x$ has at least one of elementary vague attribute values to certain degree that is bigger than 0.
\item {\bf Axiom II.} $\mathcal{M}_{x}(\bot) = 0, \mathcal{M}_{x}(\top) = 1$. \\
Axiom II says that any element $x$ must have the attribute $\varphi_{C}$.
\item {\bf Axiom III.} For any vague attribute value $A \in \Sigma_{\varphi_{C}}$, $\mathcal{M}_{x}(\neg A) = \big(\mathcal{M}_{x}(A)\big)^{N}$, where $N$ is a strong negation on $[0, 1]$.\\
Axiom III says that the degree to which $x$ doesn't has the vague attribute value $A$ is equal to the result of the strong negation operation on the degree to which it has the vague attribute value $A$.
\item {\bf Axiom IV.} The countable sum and countable product: for any vague attribute values sequence: $A_{1}, A_{2}, \cdots, A_{n}, \cdots \in \Sigma_{\varphi_{C}}$, $$\mathcal{M}_{x}\big(\veebar_{n = 1}^{\infty}A_{n}\big) = \mathcal{M}_{x}(A_{1}) \oplus \mathcal{M}_{x}(A_{2}) \oplus \cdots \oplus \mathcal{M}_{x}(A_{n}) \oplus \cdots,$$ and $$\mathcal{M}_{x}\big(\barwedge_{n = 1}^{\infty}A_{n}\big) = \mathcal{M}_{x}(A_{1}) \otimes \mathcal{M}_{x}(A_{2}) \otimes \cdots \otimes \mathcal{M}_{x}(A_{n}) \otimes \cdots,$$
    where $\oplus, \otimes$ are triangular conorm and triangular norm respectively, and they are mutually $N$-dual to each other, and $N$ is same with that of Axiom III. \\
  Axiom IV says that the degree to which $x$ has the vague attribute value $\veebar_{n = 1}^{\infty}A_{n}$ equals the result of the t-conorm operation on all these degrees to which it has each vague attribute value in the vague attribute values sequence, and that the degree to which $x$ has the vague attribute value $\barwedge_{n = 1}^{\infty}A_{n}$ equals the result of the t-norm operation on all these degrees to which it has each vague attribute value in the vague attribute values sequence.
\item {\bf Axiom V.} For any $p \in \Omega_{\varphi_{C}}$, $0 < \mathcal{M}_{x}(p) + \bigoplus_{q \in \Omega_{\varphi_{C}} \setminus \{p\}}\mathcal{M}_{x}(q) \leqslant 1$,  where $\oplus$ is a triangular conorm. \\
Axiom V says that the elements of $\Omega_{\varphi_{C}}$ are mutually exclusive to some extent. It follows that if there is $p_{0} \in \Omega_{\varphi_{C}}$ such that $\mathcal{M}_{x}(p_{0}) = 1$, then for any $p \in \Omega_{\varphi_{C}}, p \neq p_{0}$, we have $\mathcal{M}_{x}(p) = 0$. In other words, if $x$ has an elementary vague attribute value $p$  to the degree 1, then $x$ will not has other any elementary vague attribute value except $p$. In this sense, $\Omega_{\varphi_{C}}$ makes a "partition" of $U_{\varphi_{C}}$, of course, this kind of partition is not a partition in classical sense, and is a vague partition mentioned in Section 3.
\end{itemize}
\par The vague membership space $(\Omega_{\varphi_{C}}, \Sigma_{\varphi_{C}}, \mathcal{M}_{x}, \mathcal{F}, \mathcal{T})$ is said to be regular if we take $\mathcal{M}_{x}(p) + \bigoplus_{q \in \Omega_{\varphi_{C}} \setminus \{p\}}\mathcal{M}_{x}(q) = 1$ in Axiom V.
\par From the definition of vague space and Axiom II-IV, it is easy to find that the mapping $\mathcal{M}_{x}$ can be determined only by its values on $\Omega_{\varphi_{C}}$. On the other hand, the system of Axiom I-V are consistent, which can be shown by the following example.
\begin{example} Let $U_{\varphi_{C}} = [0, 200]$ and $$\Omega_{\varphi_{C}} = \{[0, 40], (40, 80], (80, 120], (120, 160], (160, 200]\},$$ $\bot = \emptyset$ and $\top = [0, 200]$, $\veebar, \barwedge, \neg$ represent the basic set operations, namely union, intersection and complement with respect to $U_{\varphi_{C}}$, respectively. $\Sigma_{\varphi_{C}}$ is the set field generated by $\Omega_{\varphi_{C}}$. $N$ is the standard negation on $[0, 1]$, $\oplus$ is the maximum or G\"{o}del t-conorm and $\otimes$ is the minimum or G\"{o}del t-norm. Let $x = 25$, define the function $\mathcal{M}_{x}$ as follows: $\mathcal{M}_{x}(\bot) = 0$, $\mathcal{M}_{x}(\top) = 1$, $\mathcal{M}_{x}([0, 40]) = 1$ and $$\mathcal{M}_{x}((40, 80]) = \mathcal{M}_{x}((80, 120]) =\mathcal{M}_{x}((120, 160]) = \mathcal{M}_{x}((160, 200]) = 0.$$ It is easy to show that $\mathcal{M}_{x}$ is a vague membership measure defined on the vague space $(\Omega_{\varphi_{C}}, \Sigma_{\varphi_{C}}, \mathcal{F})$. Hence, $(\Omega_{\varphi_{C}}, \Sigma_{\varphi_{C}}, \mathcal{M}_{x}, \mathcal{F}, \mathcal{T})$ is a regular vague membership space.
\end{example}
\par A vague membership space $(\Omega_{\varphi_{C}}, \Sigma_{\varphi_{C}}, \mathcal{M}_{x}, \mathcal{F}, \mathcal{T})$ is said to be normal if there is an element $p_{0} \in \Omega_{\varphi_{C}}$ such that $\mathcal{M}_{x}(p_{0}) = 1.$  In this case, $x$ is said to be crisp in the vague membership space $(\Omega_{\varphi_{C}}, \Sigma_{\varphi_{C}}, \mathcal{M}_{x}, \mathcal{F}, \mathcal{T})$. In fact, any partition in classical sense of set $X$ is a normal vague membership space with respect to any $x \in X$, as it was shown in the above example. It is easy to show that any normal vague membership space must be regular, not vice versa.
\begin{example}
Let $C$ be the concept  "\textit{Man}" and $\varphi_{C}$ the attribute "\textit{Age}", $U_{\varphi_{C}} = [0, 150]$ and $\Omega_{\varphi_{C}} = \{youny, old\}$,  $\bot =$ "\textit{no age}" and $\top =$ "\textit{age}". $\veebar, \barwedge, \neg$ represent the connectives "or", "and" and "negation" respectively. Define $$\Sigma_{\varphi_{C}} = \{no \ age, age, young, old, not \ age, not \ old, young \ or \ old, \cdots \}$$ such that $(\Omega_{\varphi_{C}}, \Sigma_{\varphi_{C}}, \{\veebar, \barwedge, \neg\})$ is a vague space with respect to "\textit{Age}", the operations $N$, $\oplus$ and $\otimes$ are defined as in Example 4.4. Let $x = 35$, define the function $\mathcal{M}_{x}$ as follows: $\mathcal{M}_{x}(no \ age) = 0$, $\mathcal{M}_{x}(age) = 1$, $\mathcal{M}_{x}(young) = \mathcal{M}_{x}(not \ old) = 0.6$, $\mathcal{M}_{x}(old) = \mathcal{M}_{x}(not \ young) = 0.4$, $\mathcal{M}_{x}(young \ or \ old) = \max\{0.6, 0.4\} = 0.6$, $\cdots \cdots$. It is easy to show that $\mathcal{M}_{x}$ is a vague membership measure defined on the vague space $(\Omega_{\varphi_{C}}, \Sigma_{\varphi_{C}}, \mathcal{F})$. Hence, $(\Omega_{\varphi_{C}}, \Sigma_{\varphi_{C}}, \mathcal{M}_{x}, \mathcal{F}, \mathcal{T})$ is a regular vague membership space, but not a normal vague membership space.
\end{example}
\par In the following, some basic notions will be defined so that we can characterize vague space and vague membership space more specifically. Let $a_{x} = \bigoplus_{p \in \Omega_{\varphi_{C}}}\{\mathcal{M}_{x}(p)\}$ and $b = \min_{x \in U_{\varphi_{C}}}\{a_{x}\}$, then the number $a_{x}$ is called the degree of sharpness of $x$ in vague membership space $(\Omega_{\varphi_{C}}, \Sigma_{\varphi_{C}}, \mathcal{M}_{x}, \mathcal{F}, \mathcal{T})$, and the number $1 - b$ is called the degree of separation of the set $\Omega_{\varphi_{C}}$ of elementary vague attribute values.
\par For any $A, B \in \Sigma_{\varphi_{C}}$, let $c = \max_{x \in U_{\varphi_{C}}}\{\mathcal{M}_{x}(A \barwedge B)\}$, $c$ is called as the consistent degree of $A$ and $B$ in vague space $(\Omega_{\varphi_{C}}, \Sigma_{\varphi_{C}}, \mathcal{F})$; $A$ and $B$ are said to be incompatible in vague space $(\Omega_{\varphi_{C}}, \Sigma_{\varphi_{C}}, \mathcal{F})$ if for every element $x \in U_{\varphi_{C}}$, $\mathcal{M}_{x}(A \barwedge B) = 0$ in vague membership space $(\Omega_{\varphi_{C}}, \Sigma_{\varphi_{C}}, \mathcal{M}_{x}, \mathcal{F}, \mathcal{T})$.
\par The mathematical analysis of vague phenomena, as one branch of mathematics, could and should be developed from the perspective of axiomatization in exactly the same way as Probability, and Geometry and Algebra. This means that after we have defined the elements to be studied and their basic relations, and have stated the axioms by which these relations are to be governed, all further exposition must based exclusively on these axioms, independent of the usual concrete meaning of these elements and their relations.
\par In accordance with the above discussion, in Section 3 the set of vague attribute values is defined as a free algebra on the elementary set of vague attribute values. What the elements of this set represent is of no importance in the purely mathematical development of the theory of vague membership degrees.
\par In what follows, for the sake of convenience, we denote a vague membership space $(\Omega_{\varphi_{C}}, \Sigma_{\varphi_{C}}, \mathcal{M}_{x}, \mathcal{F}, \mathcal{T})$, which is with respect to the attribute $\varphi_{C}$ of the concept $C$ and associated with $x$, by $(\Omega, \Sigma, \mathcal{M}, \mathcal{F}, \mathcal{T})$ when it does not involve any concrete problems.
\section{Redefinition of the concept of fuzzy set}
In this section, we will redefine the concept of fuzzy set  based on the essence of vagueness and the proposed axioms in Section 4. Before that, we firstly need to present the mathematical definition of vague partition.
\par In what follow, unless otherwise stated, the symbols $\mathbb{N}^{+}$ denotes the set of nonzero natural numbers $\mathbb{N} \setminus \{0\}$. For any $n \in \mathbb{N}^{+}$, the set $\{1, 2, \cdots, n\}$ be denoted by $\overline{n}$.
\begin{definition}
Let $U = [a, b] \subset \mathbb{R}$. A vague partition of $U$ is an object having the following form $$\widetilde{U} = \{\mu_{A_{1}}(x), \cdots, \mu_{A_{n}}(x)\}, n \in \mathbb{N}^{+},$$ where the functions $\mu_{A_{i}}: U \rightarrow [0, 1]$ ($i = 1, \cdots, n$) define the degrees of memberships of the element $x \in U$ to the class $A_{i}$, respectively, and satisfy the following conditions:
\par (1) for any $x \in U$, there is at least one $i \in \overline{n}$ such that $\mu_{A_{i}}(x) > 0$;
\par (2) for any $i \in \overline{n}$, $\mu_{A_{i}}(x)$ is continuous on $U$;
\par (3) for any $i \in \overline{n}$, there is at least one $x_{0} \in U$ such that $\mu_{A_{i}}(x_{0}) = 1$;
\par (4) for any $i \in \overline{n}$, if $\mu_{A_{i}}(x_{0}) = 1$ for $x_{0} \in U$, then $\mu_{A_{i}}(x)$ is non-decreasing on $[a, x_{0}]$, and is non-increasing on $[x_{0}, b]$;
\par (5) $0 < \mu_{A_{1}}(x) + \cdots + \mu_{A_{n}}(x) \leqslant 1$ holds for any $x \in U$.
\par If $\mu_{A_{1}}(x) + \cdots + \mu_{A_{n}}(x) = 1$ in (5), then $\widetilde{U}$ is said to be a regular vague partition.
\end{definition}
\par From Definition 5.1, it is easy to show that for any $i \in \overline{n}$ and any $x \in U$, $\mu_{A_{i}}(x)$ satisfies Axiom I and Axiom V, and the following two propositions are obvious.
\begin{proposition}
Let $U = [a, b] \subset \mathbb{R}$ and $\widetilde{U}  = \{\mu_{A_{1}}(x), \cdots, \mu_{A_{n}}(x)\}, n \in \mathbb{N}^{+},$ a vague partition of $U$. Then for every $x \in U$ and $i \in \overline{n}$, $$0 < \mu_{A_{i}}(x) + \max\{\mu_{A_{j}}(x) \mid j \in \overline{n}, j \neq i\} \leqslant 1.$$
\end{proposition}
\begin{proposition}
Let $U = [a, b] \subset \mathbb{R}$ and $\widetilde{U}  = \{\mu_{A_{1}}(x), \cdots, \mu_{A_{n}}(x)\}, n \in \mathbb{N}^{+},$ a vague partition of $U$. For any $i \in \overline{n}$, if $\mu_{A_{i}}(x) = 1$, then for any $j \in \overline{n}, j \neq i$, we have $\mu_{A_{j}}(x) = 0$.
\end{proposition}
\par Based on the essence of vagueness and vague partition, the concept of fuzzy set can be redefined as follows:
\begin{definition}
Let $U = [a, b] \subset \mathbb{R}$ and $\widetilde{U} = \{\mu_{A_{1}}(x), \cdots, \mu_{A_{n}}(x)\}, n \in \mathbb{N}^{+},$ a vague partition of $U$. A fuzzy set $A$ in $U$ with respect to $\widetilde{U}$ is a set of ordered pairs: $$A = \{(x, \mu_{A}(x)) \mid x \in U\},$$ where the function $\mu_{A}(x)$ can be obtained by one of the following ways within countable steps:
\par (1) there exists $i \in \overline{n}$ such that $\mu_{A}(x) = \mu_{A_{i}}(x)$ for all $x \in U$;
\par (2) $\mu_{A}(x) = \underline{\mu}(x) = 0$ for all $x \in U$;
\par (3) $\mu_{A}(x) = \overline{\mu}(x) = 1$ for all $x \in U$;
\par (4) there exists $i \in \overline{n}$ such that $\mu_{A}(x) = (\mu_{A_{i}}(x))^{N}$ for all $x \in U$, where $N$ is a strong negation on $[0, 1]$;
\par (5) there exist $i, j \in \overline{n}$ such that $\mu_{A}(x) = \mu_{A_{i}}(x) \otimes \mu_{A_{j}}(x)$ for all $x \in U$, where $\otimes$ is a triangular norm;
\par (6) there exist $i, j \in \overline{n}$ such that $\mu_{A}(x) = \mu_{A_{i}}(x) \oplus \mu_{A_{j}}(x)$ for all $x \in U$, where $\oplus$ is a triangular conorm.
\par The set of all fuzzy sets in $U$ with respect to $\widetilde{U}$ will be denoted by $\mathcal{F}(\widetilde{U})$. In fact, $\mathcal{F}(\widetilde{U})$ can be considered as a function space based on $\widetilde{U}$.
\end{definition}
\begin{remark}
In Definition 5.2, a fuzzy set is always defined on a real number interval, this is due to the fact that vagueness stems from continuity as we discussed in Section 1.
\end{remark}
\begin{remark}
It is easy to show that a fuzzy set defined by Definition 5.2 is also a fuzzy set in Zadeh's sense, but not vice versa. This partially answered the first question in Section 1, not any mapping from nonempty set $U$ to $[0,1]$ will determine a fuzzy set.
\par From Definition 5.2, on the one hand, when we define the membership function of extension of a vague concept, by vague partition $\widetilde{U}$, the relationship between this vague concept and its related vague concepts could be reflected into the definitions of its membership function. This partially answered the second question in Section 1.
\par On the other hand, it's different from Zadeh's fuzzy sets that under Definition 5.2, the basic set operations, complement, intersection and union can be defined only for these fuzzy sets under the same vague partition $\widetilde{U}$, that is, these fuzzy sets in $\mathcal{F}(\widetilde{U})$. Hence, fuzzy sets in the same domain of discourse $U$ will not always be truth-functional. In this sense, the third question in Section 1, Edgington's confusion will disappear.
\end{remark}
\begin{example}
In Netherlands in 2006, consider the concept "Man" and one of its attribute "Height", let $\Omega_{\text{Height}} = \{\text{short, medium height, tall}\}$. Let $U = [0, 3]$, and denote the vague predicate "medium height" by $A_{2}$, the membership function $\mu_{A_{2}}$ of $A_{2}$ is defined as follows
\[
\mu_{A_{2}}(x) = \left\{
               \begin{array}{ll}
                 0, & \mbox{$0 \leqslant x < 1.35$;} \\
                 2.5(x-1.35), & \mbox{$1.35 \leqslant x < 1.75$;} \\
                 1, & \mbox{$1.75 \leqslant x < 1.89$;} \\
                 20(1.94-x), & \mbox{$1.89 \leqslant x < 1.94$;} \\
                 0, & \mbox{$3 \geqslant x \geqslant 1.94$.}
               \end{array}
             \right.
\]
Denote the vague predicate "short" by $A_{1}$, and the membership function $\mu_{A_{1}}(x)$ of $A_{1}$ is defined as follows
\[
\mu_{A_{1}}(x) = \left\{
               \begin{array}{ll}
                 1, & \mbox{$0 \leqslant x \leqslant 1.35$;} \\
                 2.5(1.75-x), & \mbox{$1.35 < x \leqslant 1.75$;} \\
                 0, & \mbox{$3 \geqslant x > 1.75$.}
               \end{array}
             \right.
\]
Denote the vague predicate "tall" by $A_{3}$, and the membership function $\mu_{A_{3}}(x)$ of $A_{3}$ is defined as follows
\[
\mu_{A_{3}}(x) = \left\{
               \begin{array}{ll}
                 0, & \mbox{$0 \leqslant x < 1.89$;} \\
                 20(x-1.89), & \mbox{$1.89 \leqslant x \leqslant 1.94$;} \\
                 1, & \mbox{$3 \geqslant x > 1.94$.}
               \end{array}
             \right.
\]
It is easy to prove that $\{\mu_{A_{1}}(x), \mu_{A_{2}}(x), \mu_{A_{3}}(x)\}$ is a vague partition of $U$. By Definition 5.2, $A_{1} = \{(x, \mu_{A_{1}}(x)) \mid x \in U\}, A_{2} = \{(x, \mu_{A_{2}}(x)) \mid x \in U\}, A_{3} = \{(x, \mu_{A_{3}}(x)) \mid x \in U\}, \neg A_{1} = \{(x, (\mu_{A_{1}}(x))^{N}) \mid x \in U\}, A_{1} \barwedge A_{2} = \{(x, \mu_{A_{1}}(x) \otimes \mu_{A_{2}}(x)) \mid x \in U\}, \cdots \cdots$ are fuzzy sets in $U$.
\par Let $x = 1.5$, we have $$\mu_{A_{1}}(x)(1.5) = 0.625, \mu_{A_{2}}(x)(1.5) = 0.375, \mu_{A_{3}}(x)(1.5) = 0.$$ Obviously, compared with Zadeh's fuzzy sets, here the membership degrees $0.625$ with respect to $A_{1}$, $0.325$ with respect to $A_{2}$ and $0$ with respect to $A_{3}$ could provide us with more comprehensive information about the height attribute of the object with height of $1.5$.
\par Conversely, given membership degree (e.g. $0.4$) with respect to the fuzzy set $A_{2}$, how to find the object $x$ whose membership degree to $A_{2}$ is exactly equal to $0.4$? For this, if we also know other relevant information (e.g. the membership degree $0.6$ with respect to $A_{3}$, the membership degree $0$ with respect to $A_{1}$), that is, we know that $\mu_{A_{1}}(x) = 0, \mu_{A_{2}}(x) = 0.4, \mu_{A_{3}}(x) = 0.6$, then from these membership degrees and the membership functions of $A_{1}$, $A_{2}$ and $A_{3}$, it is easy to find that $x = 1.92$.
\end{example}
From the above example, it is easy to see that compared with Zadeh's fuzzy sets, the fuzzy sets by Definition 5.2 could describe vague concepts more delicately. This is easy to understand because Definition 5.2 includes the background information (the vague partition) of a vague concept into the definition of its membership function.
\par In addition, it's worth pointing out that intuitionistic fuzzy sets \cite{Atan86} are similar to fuzzy sets by Definition 5.2 in some ways.
\par Intuitionistic fuzzy sets model the extensions of vague concepts by two functions, the membership function and the nonmembership function. By the mutual restriction between the two functions, the subjectivity existing in defining the membership function will be weakened, and then more reasonable membership function can be obtained. However, intuitionistic fuzzy sets are still fail to reflect the relationship between a vague concept and its related vague concepts, which can be found from the following example:
\par Consider vague concept "\textit{men of medium height}" (denoted by $A_{2}$) in Netherlands in 2006, its membership function is $\mu_{A_{2}}(x)$ as defined in Example 5.1, its nonmembership function can be defined as $\nu_{A_{2}}(x) = 1 - \mu_{A_{2}}(x)$. Let Tom and Jack are two Dutch men, Tom is 1.51 metres tall, Jack is 1.92 metres tall. By the intuitionistic fuzzy set $$A_{2} = \{(x, \mu_{A_{2}}(x), \nu_{A_{2}}(x)) | x \in U\},$$ we have $\mu_{A_{2}}(1.51) = \mu_{A_{2}}(1.92) = 0.4$ and $\nu_{A_{2}}(1.51) = \nu_{A_{2}}(1.92) = 0.6$. However, Tom and Jack are two totally different men in height, but this kind of difference can't be reflected by the corresponding intuitionistic fuzzy set $A_{2}$.

\section{Conclusion}
In this paper, we explored the essence and features of vague phenomena, proposed an axiomatic system to govern membership degrees and their interconnections. More importantly, we refined the concept of fuzzy set based on vague partition from the perspective of axiomatization.
\par The approach used in the paper is similar to the axiomatization of probability theory by Kolmogorov \cite{Kolmogorov1933} in some ways. This is done under the consideration that although vagueness is different from randomness from their origins as we have mentioned in Section 1, but they are similar in other ways, especially from the perspective of epistemology. Probability of a random event is a numerical measure of the likeliness that this event will occur. Probability theory focus more on the relationship among random events in a sample space, rather than the accuracy of the probability values (numbers between 0 and 1). In other words, probability theory characterizes random phenomena from a global and overall point of view (consider random events in a probability space). The mathematical model treating vagueness aims at modelling vague phenomena approximately in terms of a kind of numerical measure, and the key point we care about is also not the accuracy (e.g. 0.6 or 0.59) of degrees of membership of objects in fuzzy sets, but is the consistency between vague predicates and their membership functions. In other words, the main aim of fuzzy set theory is to provide a numerical method to describe the relationship among objects and the relationship among vague predicates, thus membership functions of fuzzy sets need to be able to reflect definitely these relationships. Hence, the relationship among vague attribute values should also be the starting point to recognize and model vague phenomena from a quantitative view. In this sense, probability theory and vague membership theory are similar.

\par Meanwhile, we should also be aware that Lebesgue's theories of measure and integration, which have been applied successfully to establish theory of probability, are not suitable to establish the axiomatical foundation of membership degree theory for vagueness. For example, the intersection of two sets ($\{medium\ height\}$ and $\{tall\}$) is an empty set, but it is obvious that the conjunction of two vague attribute values "\textit{medium height}" and "\textit{tall}" can generated a new vague attribute value. Hence, in this paper, we established the axiomatical foundation of membership degree theory based on free algebra not sigma algebra.
\par In addition, this paper also explained the relation between natural language and fuzzy sets, that is, natural language provides a kind of tool to discretize a continual and evolutionary process. Another point we hope to emphasize here is that fuzzy sets discussed in this paper are always defined on a real number interval, we insist that fuzzy sets defined on finite universe have nothing to do with vague phenomena from the essence.
\par We hope that the work in this paper should provide with an axiomatical mathematical model for dealing with vague phenomena from a many-valued point of view. It's also our hope that this work can serve as the axiomatical foundation to expound those long-standing controversies and divergences in fuzzy set theory and its applications. We think that a good formalized theory or method treating vagueness should have the resources to accommodate all the different types of vague phenomena, and its intuitive meaning is clear.
\par Maybe you have found that in this paper, our discussion about vague concepts is proceeded in a fixed time and space. In the coming work, we will consider the time and space factors into the vague partition just like the concept of stochastic process in probability theory, we hope to obtain some corresponding results which can be used to model the vagueness of in different states, or to model the change of vague phenomena over time.
\section{Acknowledgements}
I would like to express my warm thanks to Prof. Y. Xu, P. Eklund and D.W. Pei for valuable discussions on some of the problems considered here.
\par The work was partially supported by the National Natural Science Foundation of
China (Grant No.  61305074, 61673320), the Fundamental Research Funds for the Central Universities of China (Grant No.2682014ZT28).





\bibliographystyle{model1b-num-names}
\bibliography{<your-bib-database>}



\end{spacing}

\end{document}